\pdfoutput=1

\documentclass[11pt]{article}

\usepackage{acl}

\usepackage{times}
\usepackage{latexsym}

\usepackage[T1]{fontenc}

\usepackage[utf8]{inputenc}

\usepackage{microtype}

\usepackage{graphicx}
\usepackage{subfigure} 
\usepackage{amsmath}
\usepackage{amssymb}
\usepackage{threeparttable}
\usepackage{booktabs}
\usepackage{multirow}
\usepackage{makecell}
\usepackage{comment}

\usepackage[linesnumbered,ruled,vlined]{algorithm2e}

\SetCommentSty{mycommfont}
\SetKwInput{KwInput}{Input}                
\SetKwInput{KwOutput}{Output}              

\makeatletter
\usepackage{enumitem}
\def\algbackskip{\hskip-\ALG@thistlm}
\makeatother
\usepackage{pifont}
\usepackage{xcolor}
\usepackage{tabularx}
\newcommand{\cmark}{\ding{51}}%
\newcommand{\xmark}{\ding{55}}%
\DeclareMathOperator*{\argmaxA}{arg\,max}
\DeclareMathOperator*{\argminA}{arg\,min}
\newcommand{\Lagr}{\mathcal{L}}
\usepackage{threeparttable}
\usepackage{booktabs}
\usepackage{multirow}
\usepackage[
  separate-uncertainty = true,
  multi-part-units = repeat
]{siunitx}
\usepackage{tikz}

\usepackage{array}
\newcommand\numberthis{\addtocounter{equation}{1}\tag{\theequation}}
\newcolumntype{B}{!{\hspace{-1.5ex}}c}
\newcolumntype{D}{!{\hspace{-2ex}}c}
\newcolumntype{A}{!{\hspace{-1.2ex}}l}

\title{Prompt for Extraction? PAIE: Prompting Argument Interaction for \\ Event Argument Extraction}
\author{
 Yubo~Ma$^{1\dag\ast}$, Zehao~Wang$^{2\dag\ast}$, Yixin Cao$^{\ddag3}$ \\ 
 \textbf{Mukai Li$^{4\dag}$, Meiqi Chen$^{5\dag}$, Kun~Wang$^4$, Jing~Shao$^4$} \\
 $^1$ S-Lab, Nanyang Technological University
 $^2$ KU Leuven \\
 $^3$ Singapore Management University 
 $^4$ SenseTime Research
 $^5$ Peking University\\
\texttt{yubo001@e.ntu.edu.sg, zehao.wang@esat.kuleuven.be}\\
}

\begin{document}
\maketitle

\renewcommand{\thefootnote}{\fnsymbol{footnote}}
\footnotetext[1]{Equal Contribution.}
\footnotetext[2]{Work was done when Yubo, Zehao, Mukai and Meiqi were intern researchers at SenseTime Research.}
\footnotetext[3]{Corresponding Author.}
\renewcommand{\thefootnote}{\arabic{footnote}}

\begin{abstract}
In this paper, we propose an effective yet efficient model PAIE for both sentence-level and document-level Event Argument Extraction (EAE), which also generalizes well when there is a lack of training data. On the one hand, PAIE utilizes prompt tuning for extractive objectives to take the best advantages of Pre-trained Language Models (PLMs). It introduces two span selectors based on the prompt to select start/end tokens among input texts for each role. On the other hand, it captures argument interactions via multi-role prompts and conducts joint optimization with optimal span assignments via a bipartite matching loss. Also, with a flexible prompt design, PAIE can extract multiple arguments with the same role instead of conventional heuristic threshold tuning. We have conducted extensive experiments on three benchmarks, including both sentence- and document-level EAE. The results present promising improvements from PAIE ($3.5\%$ and $2.3\%$ F1 gains in average on three benchmarks, for PAIE-base and PAIE-large respectively). Further analysis demonstrates the efficiency, generalization to few-shot settings, and effectiveness of different extractive prompt tuning strategies. Our code is available at \url{https://github.com/mayubo2333/PAIE}.
\end{abstract}

\section{Introduction}

\begin{figure}[t]
\centering
\includegraphics[width=1.0\linewidth]{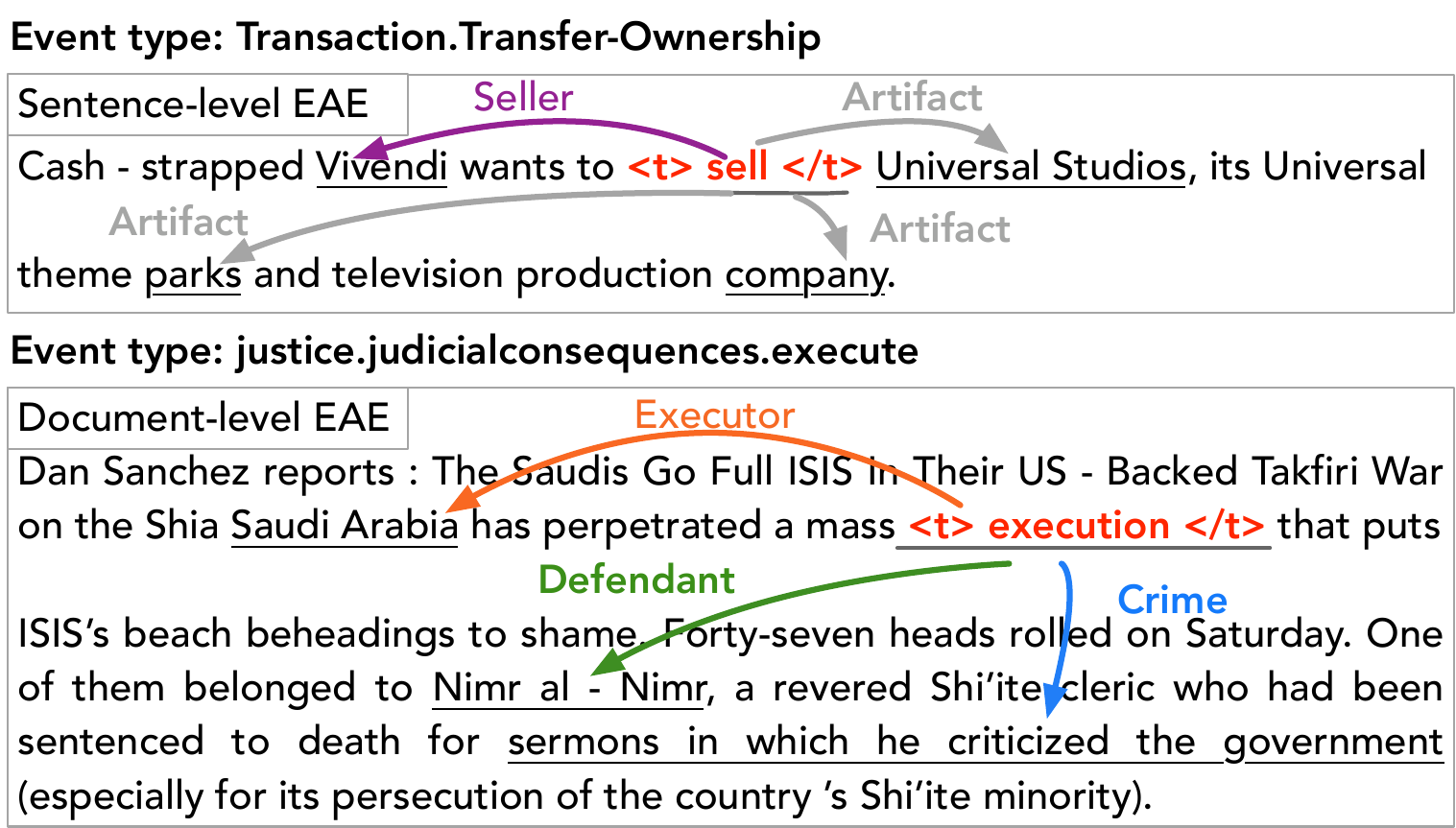}
\caption{Examples of (top) sentence-level and (bottom) document-level event argument extraction. Trigger words are included in special tokens <t> and </t>. Underlined words denote arguments and arcs denote roles.}
\label{fig: D-EAE task example}
\centering
\end{figure}

Understanding text by identifying the event and arguments has been a long-standing goal in Natural Language Processing (NLP)~\cite{sundheim-1992-overview}. As shown in Fig.~\ref{fig: D-EAE task example}, we can quickly understand that the document is talking about a \textit{Sell} event, with four involved arguments, i.e., \textit{Vivendi} (Seller), \textit{Universal Studios} (Artifact), \textit{parks} (Artifact), and \textit{company} (Artifact), where the argument roles are in brackets. Since event detection has achieved great success in recent years~\cite{wang-etal-2021-cleve}, the main challenge lies in Event Argument Extraction (EAE).

Typical efforts in EAE can be roughly classified into two groups. The first group of methods formulates it as a semantic role labeling problem~\cite{wei-etal-2021-trigger}. There are generally two steps --- first identifying candidate spans and then classifying their roles. Although joint models are proposed to optimize them together, high dependence on candidates may still suffer from error propagation~\cite{li-etal-2013-joint}. In the second group, recent studies tend to follow the success of Pre-trained Language Models (PLMs) and solve EAE by Question Answering (QA)~\cite{liu-etal-2021-machine, wei-etal-2021-trigger, du-cardie-2020-event, liu-etal-2020-event, li-etal-2020-event} and Text Generation~\cite{lu-etal-2021-text2event, li-etal-2021-document}. QA-based models can effectively recognize the boundaries of arguments with role-specific questions, while the prediction has to be one by one. 
Generation-based methods are efficient for generating all arguments, but sequential predictions degrade the performance on long-distance and more arguments. Besides, the state-of-the-art performance is still unsatisfactory (around 68\% F1 on the widely used dataset ACE05~\cite{doddington-etal-2004-automatic}). Here raises an interesting question, is there any way to combine the merits of the above methods, as well as to boost the performance?

This paper targets real scenarios, which require the EAE model to be effective yet efficient at both sentence and document levels, and even under the few-shot setting without sufficient training data. To do this, we highlight the following questions:

\begin{itemize}[leftmargin=*]
    \setlength{\parskip}{0pt}
   \setlength{\itemsep}{0pt plus 1pt}
    \item How can we extract all arguments simultaneously for efficiency?
    \item How to effectively capture argument interactions for long text, without knowing them in advance?
    \item How can we elicit more knowledge from PLMs to lower the needs of annotation?
\end{itemize}

In this paper, we investigate prompt tuning under an extractive setting and propose a novel method \textbf{PAIE} that \textbf{P}rompting \textbf{A}rgument \textbf{I}nteractions for \textbf{E}AE. It extends QA-based models to handle multiple argument extraction and meanwhile takes the best advantage of PLMs. The basic idea is to design suitable templates to prompt all argument roles for PLMs, and obtain role-specific queries to jointly select optimal spans from the text. Thus, instead of unavailable arguments, each role in the template serves as a slot for interactions, and during learning, PLMs tend to fill these slots with exact arguments via a matching loss. By predicting arguments together, PAIE enjoys an efficient and effective learning procedure. Besides, the inter-event knowledge transfer between similar role prompts alleviates the heavy burden of annotation cost.

Specifically, for prompting extraction, we design two span selectors based on role prompts, which select start/end tokens among input texts. We explore three types of prompts: manual template, concatenation template, and soft prompt. They perform well at both sentence-level EAE (S-EAE) and document-level EAE (D-EAE) and ease the requirements of the exhaustive prompt design. For joint span selection, we design a bipartite matching loss that makes the least-cost match between predictions and ground truth so that each argument will find the optimal role prompt. It can also deal with multiple arguments with the same role via flexible role prompts instead of heuristic threshold tuning.
We summarize our contributions as follow:
\begin{itemize}[leftmargin=*]
    \setlength{\parskip}{0pt}
   \setlength{\itemsep}{0pt plus 1pt}
    \item We propose a novel model, PAIE, that is effective and efficient for S-EAE and D-EAE, and robust to the few-shot setting.
    \item We formulate and investigate prompt tuning under extractive settings, with a joint selection scheme for optimal span assignments.
    \item We have conducted extensive experiments on three benchmarks. The results show a promising improvement with PAIE ($3.5\%$ and $2.3\%$ F1 gains on average absolutely in base and large model). Further ablation study demonstrates the efficiency and generalization to few-shot settings of our proposed model, as well as the effectiveness of prompt tuning for extraction. 
\end{itemize}
\section{Related Works}
\label{sec:related}

\paragraph{Event Argument Extraction:}
Event Argument Extraction is a challenging sub-task of event extraction (EE). 
There have been great numbers of studies on EAE tasks since an early stage~\cite{chen-etal-2015-event, nguyen-etal-2016-joint-event, huang-etal-2018-zero, yang-etal-2018-dcfee, Sha2018JointlyEE, zheng-etal-2019-doc2edag}. 
\citet{huang-peng-2021-DVN} propose to leverage Deep Value Networks
(DVN) that captures cross-event dependencies for EE. \citet{huang-jia-2021-SCDEE} convert documents to unweighted graph and use GAT to alleviate
the role overlapping issue. A common idea is to first identify argument candidates and then fill each with a specific role via multi-label classification~\cite{,lin-etal-2020-joint}.  To deal with implicit arguments and multiple events, \citet{xu-etal-2021-document} construct a heterogeneous graph of arguments, while DEFNN \cite{yang-etal-2021-document} predicts arguments via Parallel Prediction Networks.

A recent trend formulates EAE as an extractive question answering (QA) problem~\cite{du-cardie-2020-event, liu-etal-2020-event}. This paradigm naturally induces the language knowledge from pre-trained language models by converting EAE tasks to fully-explored reading comprehension tasks via a question template. \citet{wei-etal-2021-trigger} considers the implicit interaction among roles by adding constraints with each other in template, while~\citet{liu-etal-2021-machine} leverages data augmentation to improve the performance. However, they can only predict roles one by one, which is inefficient and usually leads to sub-optimal performance.

With the help of the pre-trained Encoder-Decoder Transformer architecture, such as BART~\cite{lewis-etal-2020-bart} and T5~\cite{2020t5}, there are also some recent works converting extraction tasks to generation tasks. \citet{paolini2021structured} propose TANL to handle a variety of structured prediction tasks, including EAE, by a unified text-to-text approach and extract all arguments in a single pass. \citet{lu-etal-2021-text2event} follow TANL and also take EAE as a sequential generation problem. \citet{li-etal-2021-document} target generation model by designing specific templates for each event type. In comparison, we prompt argument interactions to guide PLMs and optimize the multiple argument detection by designing a bipartite matching loss. This not only improves the understanding of long-distance argument dependencies but also enjoys an efficient procedure via prompt-based learning.

\paragraph{Prompt-based Learning:}
Prompt-based learning is a new paradigm emerging in the field of pre-trained language models \cite {liu2021pretrain}. Unlike the pre-training and fine-tuning paradigm, prompt-based methods convert the downstream tasks to the form more consistent with the model's pre-training tasks. \citet{schick-schutze-2021-exploiting} convert a variety of classification problems to cloze tasks by constructing related prompts with blanks and finding a mapping from particular filled words to predicted categories. \citet{li-liang-2021-prefix} focus on generation tasks and propose lightweight prefix tuning by freezing model parameters and only adjusting a sequence of continuous task-specific vectors. Different from the above prompt tuning methods designed for classification or generation tasks, our proposed method returns to \textbf{linear head} setting for fitting extraction task better. It is somewhat similar as a concurrent work P-tuning v2~\cite{DBLP:journals/corr/abs-2110-07602}. 
\begin{figure*}[t]
    \centerline{\includegraphics[width=0.9\linewidth]{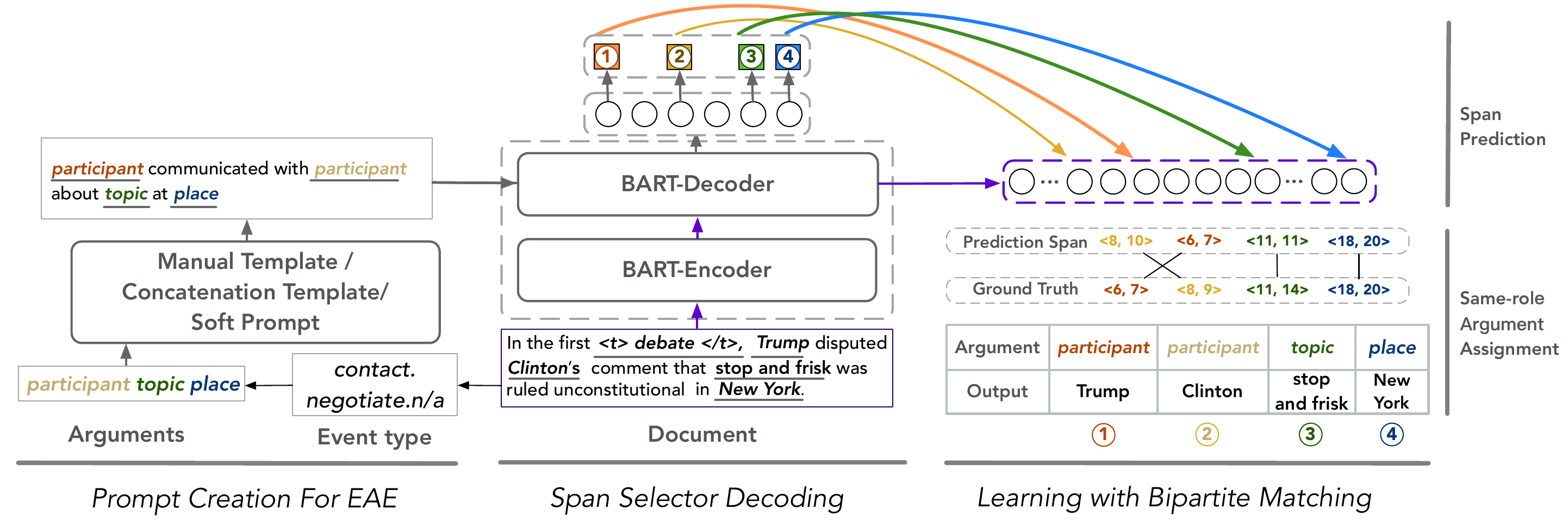}}
    \caption{The overall architecture of PAIE. Given a context (about an event), PAIE first creates joint prompts based on its event type. Then the context and prompt are fed into the BART-Encoder and BART-Decoder to generate context representation and role-specific span selectors. Multiple span selectors extract argument spans from the context simultaneously. A bipartite matching loss finally optimizes the global span assignment.}

    \label{fig: model framework}
\end{figure*}

\section{Methodology}
\label{sec:method}

PAIE considers multiple arguments and their interactions to prompt PLMs for joint extraction. Our model, as illustrated in Fig.~\ref{fig: model framework}, contains three core components: \textit{prompt creation}, 
\textit{span selector decoding}, 
and \textit{span prediction}. In the following sections, we will first formulate prompt for extraction, and describe each component in turn.

\subsection{Formulating Prompt for Extraction}
\label{subsec: formulation}
Existing prompt-based methods mainly focus on classification and generation tasks. Conventional extraction objectives are converted into a generation task. This brings an inefficiency issue that the model has to enumerate all of extraction candidates. For example, \citet{cui-etal-2021-template} design the prompt for named entity recognition: \textit{[candidate span] is [entity type/not a] entity}. The models need to fill the first slot with candidate entities, and check the outputs of LM for the second slot for extraction. Can prompt-based methods directly be applied on extraction? since the basic idea is similar with classification/generalization --- comparing the slot embeddings with label vocabulary/input tokens. Here, we give a formulation about the general extractive prompting method and then apply it on EAE for a case study.

\noindent{\textbf{(1)}} \textit{Prompt Creation}. Given context $X$ and a series of queries $Q = \{q_1, q_2, ..., q_K\}$, we create a joint prompt containing all these queries, where $f_{prompt}$ is the prompt creator.
$$
Pt = f_{prompt}(Q)
$$

\noindent{\textbf{(2)}} \textit{Prompted Selector Decoding}. 
Given a PLM $\Lagr$, context $X$, and prompt $Pt$, we decode a query-specific (answering) span selector as follows:
$$
\theta_{q_k} = h_{\Lagr}(q_k; Pt, X) 
$$
where $q_k$ is the $k$-th query in the prompt and $h_{\Lagr}$ is the outputs of PLMs.

\noindent{\textbf{(3)}} \textit{Prompted Span Selection}. To find the optimal span, we design two selectors for the start and end tokens from context: 
$$
(s, e)_{q_k} = \text{Span-search} [g_{\Lagr}(X; \theta_q)]
$$
where $(s, e)_{q_k}$ is the span about $k$-th query and $g_{\Lagr}$ is the span selector. Clearly, such formulation is better than generative extraction by mainly considering the adjacent constraints of span.

\paragraph{Task Definition}
We formulate EAE task as a prompt-based span extraction problem on dataset $D$. Given an instance $(X, t, e, R^{(e)}) \in D$, where $X$ denotes the context, $t \subseteq X$ denotes the trigger word, $e$ denotes the event type and $R^{(e)}$ denotes the set of event-specific role types, we aim to extract a set of span $A$. Each $a^{(r)} \in A$ is a segmentation of $X$ and represents an argument about $r \in R^{(e)}$.

\subsection{Prompt Creation for EAE}
\label{subsec: joint prompt creation}
We create a set of prompts for each event type $e$ in dataset $D$. Each prompt contains all roles $r \in R^{(e)}$. For example in Fig.\ref{fig: model framework}, given event type $e$ as \textit{negotiate} and $R^{(e)}$ as $\{\textit{Participant}, \textit{Topic}, \textit{Place}\}$, the prompt $Pt^{(e)}$ may be defined as follows:

\textit{\underline{Participant} communicated with \underline{Participant} about \underline{Topic} at \underline{Place} .}

We call the mentions of roles in the prompt as \textbf{slot}, and there are four slots underlined in this example (and colored in Fig.~\ref{fig: model framework}). Such design allows our model to capture the implicit interactions among different roles.

To avoid threshold tuning for multiple arguments with the same role, 
the prompt is flexible to use multiple slots for the same role, such as role \textit{Participant} in the above example. The number of slots for the role is heuristically determined according to the maximum number of arguments of each role in the training dataset. We design three different prompt creators $f_{prompt}$, the mapping from a set of roles to a prompt as follows:
\begin{enumerate}[leftmargin=*]
    \setlength{\parskip}{0pt}
   \setlength{\itemsep}{0pt plus 1pt}
    \item Manual Template: All roles are connected manually with natural language. We follow the template from~\citet{li-etal-2021-document} for fair comparison.
    \item Soft Prompt: Following~\citet{qin-eisner-2021-learning} and \citet{liu2021gpt}, we connect different roles with learnable, role-specific pseudo tokens.
    \item Concatenation Template: To concatenate all role names belonging to one event type.
\end{enumerate}

\noindent{We give one example of these three types of prompt in Table~\ref{tab: diff prompt} and list more examples in Appendix~\ref{subsec: prompt example}. Further analysis can be found in Section~\ref{subsec: prompt variants}.}

\begin{table*}[t]
\small
    \centering
    \begin{threeparttable}
    \begin{tabular}{c|c}
    \toprule
    \textbf{Prompt Type} & \textbf{Prompt Example} \\
    \midrule
    MA Template  & \makecell{\underline{Victor} ( and \underline{Victor} ) defeated in \underline{ConflictOrElection} at \underline{Place}}( and \underline{Place} ) \\
    \midrule
    SF Prompt & \makecell{<Vic\_left0> \underline{Victor} <Vic\_right0> ( <Vic\_left0> \underline{Victor} <Vic\_right0> ) \\ <Conf\_left0> \underline{ConflictOrElection} <Conf\_right0> \\ <Place\_left0> \underline{Place} <Place\_right0> ( <Place\_left0> \underline{Place} <Place\_right0> )} \\
    \midrule
    CA Template  & \underline{Victor} ( \underline{Victor} ) \underline{ConflictOrElection} \underline{Place} ( \underline{Place} )\\
    \bottomrule
    \end{tabular}
    \end{threeparttable}
    \caption{Variants of prompt introduced in section~\ref{subsec: joint prompt creation}.  \textbf{MA}:Manual Template. \textbf{SF}:Soft Prompt. \textbf{CA}:Concatenation Template. Words with angle brackets in Soft Prompt denote role-specific pseudo tokens of continuous prompts. For multi-argument cases, we simply add slots within square brackets.}
    \label{tab: diff prompt}
\end{table*}

\subsection{Role-specific Selector Generation}
Given context $X$ and prompt $Pt$, this module generates the role-specific span selector $\theta_k$, for each slot $k$ of the prompt. Here we choose $\Lagr$ as BART~\cite{lewis-etal-2020-bart}, a standard Transformer-based pre-trained language model consisting both an \textbf{Encoder} and a \textbf{Decoder}: $\Lagr=[\Lagr_{enc}, \Lagr_{dec}]$.

We first define text markers $\textbf{\textlangle t\textrangle}$ and $\textbf{\textlangle /t\textrangle}$ as special tokens then insert them into context $X$ before and after the trigger word respectively.
$$\tilde{X} = [x_1, x_2, ...,  \textbf{\textlangle t\textrangle}, x_{trig}, \textbf{\textlangle /t\textrangle}, ... , x_n]$$

Instead of concatenating the processed context $\tilde{X}$ and prompt $Pt$ directly, we feed the context into BART-Encoder and the prompt into BART-Decoder separately, as illustrated in Fig.~\ref{fig: model framework}. The prompt and context would interact with each other at the cross-attention layers in the decoder module.

\begin{equation}
\begin{aligned}[c]
    &H_X^{(enc)} = \Lagr_{enc}(\tilde{X}) \\
    &H_X = \Lagr_{dec}(H_X^{(enc)}; H_X^{(enc)}) \\
    &H_{pt} = \Lagr_{dec}(Pt; H_X^{(enc)}) \\
\end{aligned}
\end{equation}

\noindent{where $H_X$ denotes the event-oriented context representation and $H_{pt}$ denotes context-oriented prompt representation. For $k$-th slot in the joint prompt we mean-pool its corresponding representations from $h_{pt}$ and obtain role feature $\psi_{k} \in R^h$, where $h$ denotes the dimension of hidden layer in BART. Note that a role may have multiple slots and, correspondingly, multiple role features and span selectors.}

We adopt a simple but effective modification on previous QA-based methods by deriving \textbf{role-specific span selector} $\theta_{k}$ from every role feature in the prompt. Given role feature $\psi_{k}$, we have:
\begin{align*} 
\psi_{k}^{(start)}& = \psi_{k} \circ w^{(start)} \in R^h \\ 
\psi_{k}^{(end)} &= \psi_{k} \circ w^{(end)} \in R^h
\numberthis
\end{align*}

\noindent{where $\theta = [w^{(start)}; w^{(end)}] \in R^{h \times 2}$ is learnable parameters shared among all roles, and $\circ$ represents element-wise multiplication. $\theta_{k} = [\psi_{k}^{(start)}; \psi_{k}^{(end)}]$ is exactly the span selector for $k$-th slot in the prompt. With only one meta-head $\theta$ and simple operations, our method enables to generate arbitrary number of role-specific span selectors to extract related arguments from context. Recall the generation process of role feature $\psi_{k}$ from prompt $h_{pt}$, it is obvious that both the interaction among different roles and the information aggregation between context and roles are considered under this paradigm.}

\subsection{Learning with Prompted Span Selector}
Given context representation $H_X$ and a set of span selectors $\{\theta_k\}$, each $\theta_k$ aims to extract at most one corresponding argument span $(s_k, e_k)$ from $H_X$. For $\theta_k$ relating to one argument $a_k = \tilde{X}_{i:j}$, where $i$ and $j$ are the start and end word indices in context, the selector is expected to output $(\hat{s}_k, \hat{e}_k) = (i, j)$ as prediction. And for $\theta_k$ relating to no argument (when context has no argument about this role, or the slot number of this role exceeds the argument number), it is expected to output $(\hat{s}_k, \hat{e}_k) = (0, 0)$ representing an empty argument $\epsilon$.

We first follow the extractive prompt formulation in Section~\ref{subsec: formulation} to calculate the distribution of each token being selected as the start/end of the argument for each role feature.
\begin{align*} 
\text{logit}_{k}^{(start)} &= \psi_{k}^{(start)} H_X \in R^L \\ 
\text{logit}_{k}^{(end)} &= \psi_{k}^{(end)} H_X \in R^L \numberthis
\label{eq: logit}
\end{align*}
where $\text{logit}_{k}^{(start)}$ and $\text{logit}_{k}^{(end)}$ represent start and end position distributions over the context tokens for each slot $k$, and $L$ denotes the context length.

Then we calculate probabilities where the start/end positions locate:
\begin{align*}
    p_k^{(start)} &= \text{Softmax}(\text{logit}_{k}^{(start)}) \in R^L \\
    p_k^{(end)} &= \text{Softmax}(\text{logit}_{k}^{(end)}) \in R^L 
\numberthis
\label{eq: softmax without bipartite}
\end{align*}
and define the loss function as:
\begin{align*}
 \Lagr_{k}(X) & = - (\log p_{k}^{(start)}(s_{k}) + \log p_{k}^{(end)}(e_{k})) \\
 \Lagr & = \sum_{X \in D}\sum_{k} \Lagr_{k}(X)
\label{eq: loss}
\numberthis
\end{align*}
where $D$ ranges over all context in dataset and $k$ ranges over all slots in prompt for $X$.

\noindent\textbf{Bipartite Matching}
We optionally introduce bipartite matching to deal with multiple arguments of the same role for finding the global-optimal assignments with the least-cost match. Since we insert multiple slots about this role and each slot generates one prediction, it is a canonical bipartite matching problem that matches local-optimal predictions (of each slot) and ground truth as much as possible. Following~\citet{Carion-etal-2020-detr, yang-etal-2021-document}, we use Hungarian algorithm~\cite{kuhn1955hungarian} and leave the detail about it in Appendix~\ref{ap: bipartite matching}.

\subsection{Inference}
For inference, we define the set of candidate spans for event arguments as $\mathcal{C}=\{(i, j)|(i,j) \in L^2, 0<j-i \leq l\} \cup \{(0, 0)\}$. It contains all spans shorter than the threshold $l$ \textbf{and} special span $(0, 0)$ indicating no arguments. Our model extracts the argument of each span selector $\theta_k$ by enumerating and scoring all candidate spans as:
\begin{equation}
 \text{score}_k(i, j) = \text{logit}_{k}^{(start)}(i) + \text{logit}_{k}^{(end)}(j)
\end{equation}
and the predicted span of slot $k$ is given by:
\begin{equation}
    (\hat{s_k}, \hat{e_k}) = \argmaxA_{(i, j) \in \mathcal{C}} \text{score}_k(i, j)
\end{equation}
Since at most one span is predicted by each slot in the prompt, this strategy avoids the exhaustive threshold tuning.
\section{Experiments}

In this section, we explore the following questions:

\begin{itemize}[leftmargin=*]
    \setlength{\parskip}{0pt}
   \setlength{\itemsep}{0pt plus 1pt}
    \item Can PAIE better utilize PLMs for joint extraction to boost the performance of S-EAE and D-EAE?
    \item How do different prompt training strategies affect the results?
    \item How does PAIE perform in various practical settings, including efficiency and generalization to few-shot, long-distance, and multiple arguments?
\end{itemize}

\begin{table*}
\small
    \centering
    \begin{threeparttable}
    \begin{tabular}{A A | B B B B B B B}
    \toprule
    \multirow{2}{*}{\textbf{Model}} & \multirow{2}{*}{\textbf{PLM}}  & \multicolumn{2}{c}{\textbf{ACE05}} &  \multicolumn{2}{c}{\textbf{RAMS}} & \multicolumn{3}{c}{\textbf{WIKIEVENTS}} \\
    & & \textbf{Arg-I} & \textbf{Arg-C} & \textbf{Arg-I} & \textbf{Arg-C}  & \textbf{Arg-I} & \textbf{Arg-C} & \textbf{Head-C}\\
    \midrule
    FEAE \small{~\cite{wei-etal-2021-trigger}} & BERT-b & - & - &  53.5\rlap{*} & 47.4\rlap{*} & - & - & -\\
    DocMRC \small{~\cite{liu-etal-2021-machine}} & BERT-b & - & - &  - & 45.7\rlap{*} &  - & 43.3\rlap{*} & -\\
    \multirow{2}{*}{OneIE\small{~\cite{lin-etal-2020-joint}}} & BERT-b & 65.9 & 59.2 &  - & - &  - & - & -\\
     & BERT-l & 73.2 & 69.3 &  - & - &  - & - & -\\
    \multirow{2}{*}{EEQA \small{~\cite{du-cardie-2020-event}}} & BERT-b & 68.2\rlap{*} & 65.4\rlap{*} & 46.4  &44.0 &54.3   &53.2  &56.9 \\
    & BERT-l &70.5  &68.9  & 48.7  &46.7  &56.9  &54.5 & 59.3 \\
    \multirow{2}{*}{BART-Gen\small{~\cite{li-etal-2021-document}}} & BART-b & 59.6 & 55.0 &  50.9 & 44.9 & 47.5 & 41.7  & 44.2\\
    & BART-l & 69.9\rlap{*} & 66.7\rlap{*} &  51.2 & 47.1 & 66.8 &62.4  & 65.4\\
    \multirow{2}{*}{EEQA-BART \small{(Our implementation)}} & BART-b & 69.6 & 67.7 &49.4 & 46.3 & 60.3 & 57.1 & 61.4\\
    & BART-l & 73.1 & \underline{72.2} &  51.7 & 48.7 & 61.6  & 57.4 & 61.3\\
    \midrule
    \multirow{2}{*}{PAIE \small{(Ours)}} & BART-b & \underline{73.6} & 69.8 & \underline{54.7} & \underline{49.5} & \underline{68.9} & \underline{63.4} & \underline{66.5} \\
    & BART-l & \textbf{75.7} & \textbf{72.7} &  \textbf{56.8} & \textbf{52.2} &  \textbf{70.5} & \textbf{65.3} & \textbf{68.4} \\
    \bottomrule
  \end{tabular}
  \caption{Overall performance. We highlight the best result and underline the second best. * means the value from the original paper. \textbf{b} in column \textbf{PLM} denotes base model and \textbf{l} denotes large model.}
  \label{tab: main results}
  \end{threeparttable}
\end{table*}

\subsection{Experimental Setup}
\paragraph{Datasets}
We conduct experiments on three common datasets in Event Argument Extraction task: RAMS~\cite{ebner-etal-2020-multi}, WIKIEVENTS~\cite{li-etal-2021-document} and ACE05~\cite{doddington-etal-2004-automatic}. RAMS and WIKIEVENTS are latest document-level EAE benchmarks, while ACE05 is a classical dataset commonly used for sentence-level EAE task. We leave the dataset details in Appendix~\ref{ap: dataset}.

\paragraph{Evaluation Metric}
We adopt two evaluation metrics. 
(1) Argument Identification F1 score (Arg-I): an event argument is correctly identified if its offsets and event type match those of any of the argument mentions. 
(2) Argument Classification F1 score (Arg-C): an event argument is correctly classified if its role type is also correct. For WIKIEVENTS dataset, we follow \cite{li-etal-2021-document} and additionally evaluate Argument Head F1 score (Head-C), which only concerns the matching of the headword of an argument.

\paragraph{Implementation Details} 
Please refer to Appendix~\ref{ap: implementation} for implementation details of PAIE.

\paragraph{Baselines}
We compare PAIE with several state-of-the-art models in three categories:
(1) Multi-label classification model: \textbf{ONEIE}~\cite{lin-etal-2020-joint} 
(2) Generation model: \textbf{BART-Gen}~\cite{li-etal-2021-document}
(3) QA-based model: \textbf{EEQA}~\cite{du-cardie-2020-event},  \textbf{DocMRC}~\cite{liu-etal-2021-machine} and \textbf{FEAE}~\cite{wei-etal-2021-trigger}.
For a fair comparison, we replace the PLMs used in the strongest baseline EEQA with BART, the same with PAIE, namely \textbf{EEQA-BART}. 
More details of baselines are listed in Appendix~\ref{ap: baselines}.

\begin{table*}
 \centering
  \small
    \begin{threeparttable}
        \begin{tabular}{A|BBBB|BBB} 
        \toprule
        \textbf{Model} & \multirow{2}{*}{\textbf{\shortstack{Bipartite\\Matching}}} & \multirow{2}{*}{\textbf{\shortstack{Multi-arg\\Prompt}}} & \multirow{2}{*}{\textbf{\shortstack{Role-specific\\Selector}}} & \multirow{2}{*}{\textbf{PLM}} & \multicolumn{3}{c}{\textbf{Arg-C}}  \\
        & & & & & \textbf{ACE05} & \textbf{RAMS} & \textbf{WIKI} \\
        \midrule
        \textbf{PAIE} & \cmark & \cmark & \cmark & BART-b &  $69.8${\tiny$\pm0.98$} &  $49.5${\tiny$\pm0.65$} &  $63.4${\tiny$\pm1.17$} \\
        \midrule
        \textbf{PAIE\_\tiny w/o bipartite} & \xmark & \cmark & \cmark & BART-b & $68.9${\tiny $\pm{1.03}$} & $49.4${\tiny$\pm0.98$} & $62.4${\tiny $\pm1.09$} \\
         \textbf{PAIE\_\tiny w/o multi-prompt} & \xmark & \xmark & \cmark & BART-b &  $66.9${\tiny $\pm0.61$} &  $47.6${\tiny $\pm1.20$}&  $59.9${\tiny $\pm1.26$} \\
         \textbf{EEQA-BART} & \xmark & \xmark & \xmark & BART-b &  $67.7${\tiny $\pm0.64$} &  $46.3${\tiny $\pm0.77$}&  $57.1${\tiny $\pm0.82$} \\
        \midrule
         \textbf{EEQA} & \xmark & \xmark & \xmark & BERT-b  & $65.4$ & $44.0$ &  $53.2$\\
        \bottomrule
        \end{tabular}
    \end{threeparttable}
    \caption{Ablation study on three benchmarks. WIKIEVENTS is abbreviated as WIKI (the same below). 
    }
    \label{Ablation study}
\end{table*}

\subsection{Overall Performance}
Table \ref{tab: main results} compares our approach with all baselines. We observe that PAIE performs best on all datasets. For S-EAE, our base model achieves an absolute Arg-C improvement of $2.1\%$ on ACE05. For D-EAE, our base model obtains $2.1\%$ and $6.3\%$ Arg-C gains on RAMS and WIKIEVENTS, respectively. Similarly, our large-version model achieves $3.5\%$ and $2.9\%$ gains. This demonstrates a good generalization ability of our proposed method on dealing with varying lengths of context.

We also find that QA-based model sometimes performs well even in document-level EAE tasks. The EEQA-BART model shows almost the same Arg-C with BART-Gen~\cite{li-etal-2021-document} on RAMS dataset. Other QA-based models (especially those considering interactions among arguments, like FEAE~\cite{wei-etal-2021-trigger}) also have competitive performance. As for WIKIEVENTS, however, QA-based models are inferior to sequential-generation models significantly. We speculate that the performance of previous QA-based models is not robust to handle longer text.
Both BART-Gen~\cite{li-etal-2021-document} and our model PAIE have a relatively stable performance on various document-level EAE datasets, but our model performs better, especially with smaller PLMs. 

Next, we conduct further analysis with the strongest baseline EEQA-BART and our PAIE. We use the base-version BART for a fair comparison.

\subsection{Ablation Study}
\label{subsec: ablation study}

In this section, we investigate the effectiveness of our main components by removing each module in turn. (1) \textbf{bipartite matching}. We drop out of the bipartite matching loss and ignore the global optimal span assignment. (2) \textbf{multi-arg prompt}. We additionally replace the prompt containing multiple roles with several single templates in which include only one role. (3) \textbf{role-specific selector}. The selector is not role-specific anymore but is shared among all roles. This variant degrades to EEQA-BART.

We summarize the results of ablation studies in Table~\ref{Ablation study}. \textbf{(1)} EEQA-BART outperforms EEQA significantly, which demonstrates that even conventional QA-based methods have substantial space for improvement with a better PLM and span selection strategy. \textbf{(2)} The role-specific selector further improves Arg-C scores in RAMS and WIKIEVENTS, while taking a slightly negative effect on ACE05. Since the former two datasets are document-level and have more role types (65 in RAMS, 59 in WIKIEVENTS, and 36 in ACE05), we speculate that role-specific selector plays a critical role when identifying and disambiguating roles with complicated ontology structures in long documents.  \textbf{(3)} Joint multi-argument prompt achieves consistent improvement on all three datasets. It indicates that the joint prompt has the potential to capture implicit interaction among arguments. \textbf{(4)} Bipartite matching loss has an average improvement of $0.7\%$ on three benchmarks. We conjectured it is due to the permutation-invariance property of bipartite matching and discuss further in Appendix \ref{sec: bipartite discuss}.
\section{Evaluation of Extractive Prompting}

\begin{table}[t]
 \centering
 \small
    \begin{threeparttable}
    \begin{tabular}{B B | B B B}
        \toprule
        
        \textbf{Variant} & \textbf{PLM} & \textbf{ACE05} & \textbf{RAMS} & \textbf{WIKI}  \\
        \midrule
        \multirow{3}{*}{\textbf{PAIEE}} & BE-b & 65.9 & 46.3 & 62.9 \\
         & BA-b & 70.2 & 49.3 & 62.8 \\
         & BA-l & \underline{72.3} & \underline{51.7} & \underline{65.1} \\
        \midrule
        \multirow{2}{*}{\textbf{PAIE}} & BA-b & 69.8 & 49.5 & 63.4 \\
         & BA-l &  \textbf{72.7} & \textbf{52.2} & \textbf{65.3} \\
        \bottomrule
    \end{tabular}
    \end{threeparttable}
    \caption{Arg-C F1 of different PLMs. BE and BA denote BERT and BART. Note that we also try PLM with only encoder such as BERT under PAIEE setting, which does not require a decoder.}
    \label{tab: model variation}
\end{table}

\subsection{Architecture Variants}
\label{sec: model arch}
PAIE feeds the context into BART-Encoder and the prompt into BART-Decoder respectively. A plausible and straightforward variant called \textbf{PAIEE} (PAIE-Encoder) concatenates context and prompt, then feed them into encoder directly. We investigate the performance of PAIEE compared with PAIE in this section, as shown in Table~\ref{tab: model variation}.

We can see that concatenating context and prompt slightly impairs the model performance. It seemingly indicates that the over-interaction between context and prompt is not of benefit. Furthermore, the prompt squeezes the limited input length of the encoder kept for a document if it concatenates with the document. The experiments support our strategy feeding context and prompt separately without concatenation to PAIE. 

\subsection{Prompt Variants}
\label{subsec: prompt variants}

\begin{figure*}[htbp]
\centering
\subfigure[ACE05]{
\begin{minipage}[t]{0.33\linewidth}
\centering
\includegraphics[width=\linewidth]{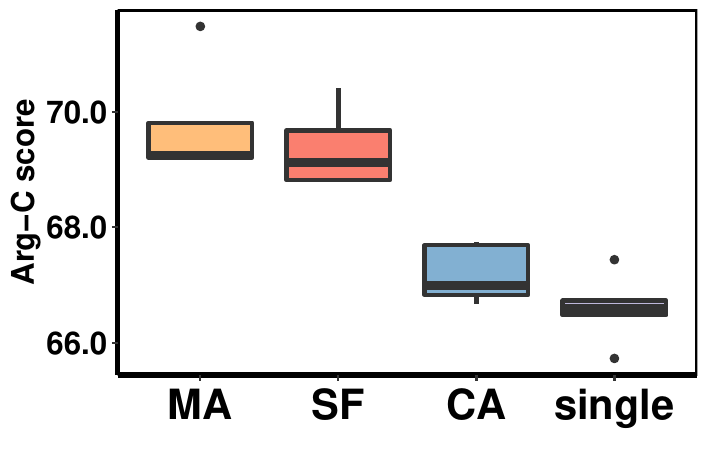}
\end{minipage}%
}%
\subfigure[RAMS]{
\begin{minipage}[t]{0.33\linewidth}
\centering
\includegraphics[width=\linewidth]{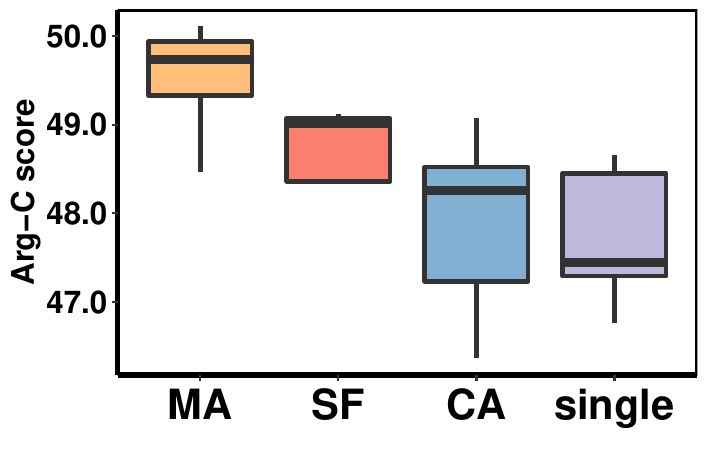}
\end{minipage}%
}%
\subfigure[WIKIEVENTS]{
\begin{minipage}[t]{0.33\linewidth}
\centering
\includegraphics[width=\linewidth]{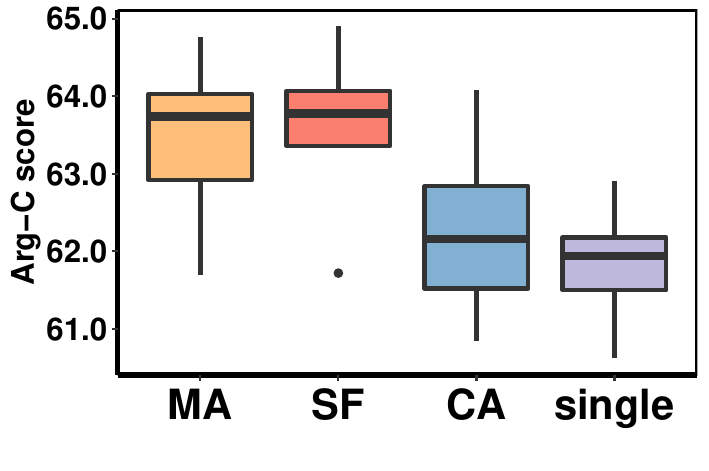}
\end{minipage}
}%
\caption{Arg-C F1 using three different types of joint prompts in Table~\ref{tab: diff prompt} plus the single template on three benchmarks. \textbf{MA}: Manual Template. \textbf{SF}: Soft Prompt. \textbf{CA}: Concatenate Template. \textbf{single}: Single Template.}
\label{fig:prompt}
\end{figure*}

We investigate how different types of prompts affect the performance in this section, as shown in Fig.~\ref{fig:prompt}. We compare four different prompts: three joint prompts introduced in Section \ref{subsec: joint prompt creation} and one single template containing only one role slot, i.e. the question template used in QA-based method.

We find that \textbf{(1)} All three joint prompts outperform the single template, which validates the effectiveness of the joint prompt. \textbf{(2)} Manual template has the most stable performance and usually the better result than others. \textbf{(3)} Soft prompt achieves comparable result with a manual template. We claim this observation inspiring because the creation of the manual template is laborious and soft prompts almost avoid such a handcrafted process. It also accords with current trends of creating distinct continuous prompts, which usually perform better than manual ones. \textbf{(4)} Concatenation template performs worst among joint prompts. We conjecture it is due to such prompt neither contains prior knowledge about role interaction (manual template) nor learns such interaction during training (soft prompt). 
\section{Analysis on Real Scenario}
\subsection{Long-range Dependencies}

\begin{table}
\centering
\small
    \begin{threeparttable}
        \begin{tabular}{D|DDDDD}
        \toprule
        \multirow{2}{*}{\textbf{Model}} & \multicolumn{5}{c}{\textbf{Trigger-Argument Distance $d$}} \\
        & $\mathbf{-2_{[79]}}$ & $\mathbf{-1_{[164]}}$ & $\mathbf{0_{[1811]}}$ & $\mathbf{1_{[87]}}$ & $\mathbf{2_{[47]}}$ \\
        \midrule
        \makecell{ \textbf{BART-Gen}} & 17.7 & 16.8 & 44.8 & 16.6 & 9.0 \\
        \makecell{ \textbf{DocMRC}} & 21.0 & 20.3 & 46.6 & 17.2 & 12.2 \\
        \makecell{\textbf{FEAE}} & \textbf{23.7} & 19.3 & 49.2 & 25.0 & 5.4 \\
        \makecell{\textbf{EEQA-BART}} & 15.6 & \underline{24.0} & 51.7 & 23.5 & 8.0 \\
        \midrule
        {\textbf{PAIE\_\tiny w/o multi-prompt}}  & 21.2 & 21.4 & \underline{52.3} & \underline{27.9} & \underline{24.6} \\
        {\textbf{PAIE}}  & \underline{21.7} & \textbf{27.3} & \textbf{54.7} & \textbf{29.4} & \textbf{25.4} \\
        \bottomrule
        \end{tabular}
    \end{threeparttable}
\caption{Performance (Arg-C F1 score) breakdown by argument-trigger distance $d$ on RAMS development set. The argument number of each case is given in the bracket. 
}
\label{tab: dist}
\end{table}

In D-EAE task, arguments could span multiple sentences. Therefore, the model is required to capture long-range dependencies. For better evaluating PAIE and comparing with others, we list their performance breakdown on different sentence distances between arguments and the given trigger word in Table~\ref{tab: dist}. 
We can see that \textbf{(1)} PAIE significantly improves the ability to extract arguments with long distances, especially for those behind the trigger words (see columns with positive $d$ values). \textbf{(2)} The last two rows of the table indicate that joint prompts in PAIE leverage the implicit interaction among roles, and roles conditioning on each other lower the difficulty to extract long-distance arguments effectively.

\subsection{Same-role Argument Assignment}
\label{subsec:same-role arg}
Multiple arguments may share the same role in the same event. We show that PAIE outperforms QA-based models dealing with it in both efficiency and effectiveness in this section.

\begin{figure}[t]
\centering
\includegraphics[width=\linewidth]{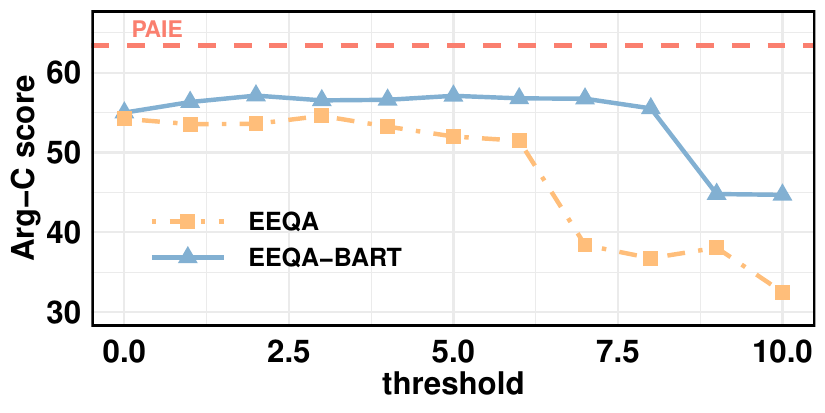}
\caption{Arg-C F1 w.r.t different thresholds for WIKIEVENTS. We draw the performance of PAIE in red dashed line for comparison (no threshold tuning).
}
\label{th_tuning}
\end{figure}

\noindent{\textbf{Efficiency}} To solve this problem, QA-based methods usually adopt the thresholding strategy, which compares the score of each text span with a manually tuned threshold. We claim that it consumes lots of time and computational resources for finding a good threshold and usually ends with sub-optimal results. We support such claim by a coarse grid search tuning span threshold on WIKIEVENTS dataset using EEQA and EEQA-BART models, as shown in Fig.~\ref{th_tuning}. The choice of threshold highly affects the performance of the model. In addition, models with the same architecture but different PLMs have totally different optimal thresholds even on the same dataset, not to mention on distinct datasets. PAIE requires no threshold tuning since each slot in the prompt only predicts at most one argument span and usually achieves much higher inference speed in practice.   

\begin{figure*}[htbp]
\centering
\subfigure[ACE05]{
\begin{minipage}[t]{0.33\linewidth}
\centering
\includegraphics[width=\linewidth]{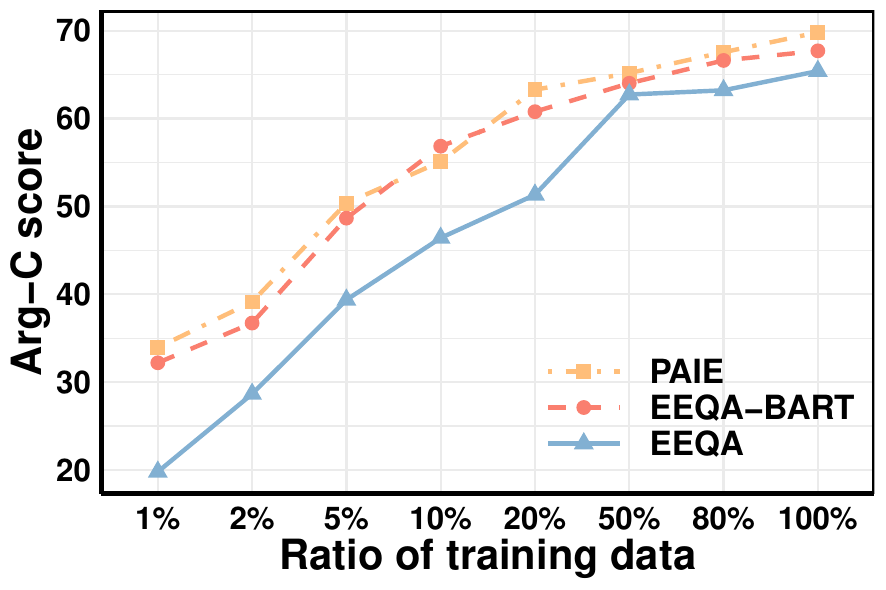}
\end{minipage}%
}%
\subfigure[RAMS]{
\begin{minipage}[t]{0.33\linewidth}
\centering
\includegraphics[width=\linewidth]{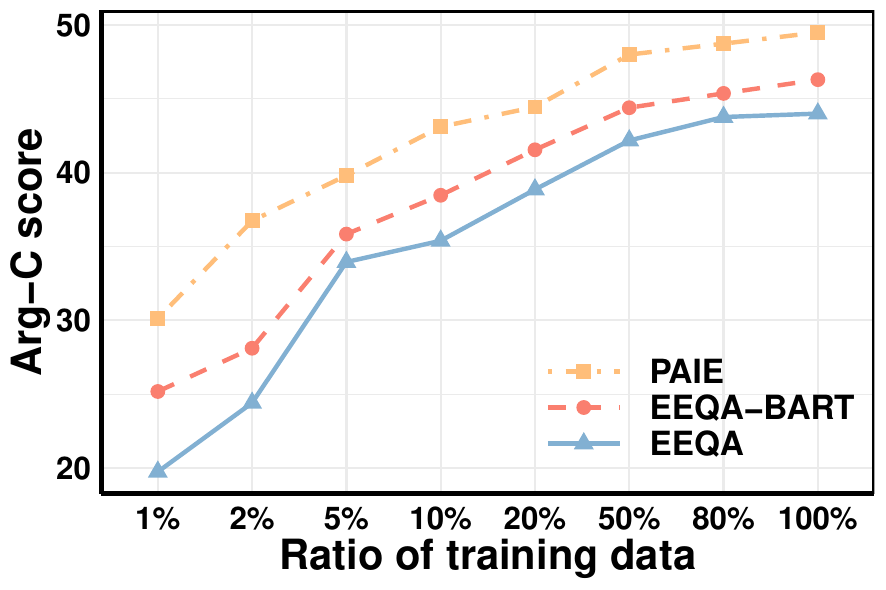}
\end{minipage}%
}%
\subfigure[WIKIEVENTS]{
\begin{minipage}[t]{0.33\linewidth}
\centering
\includegraphics[width=\linewidth]{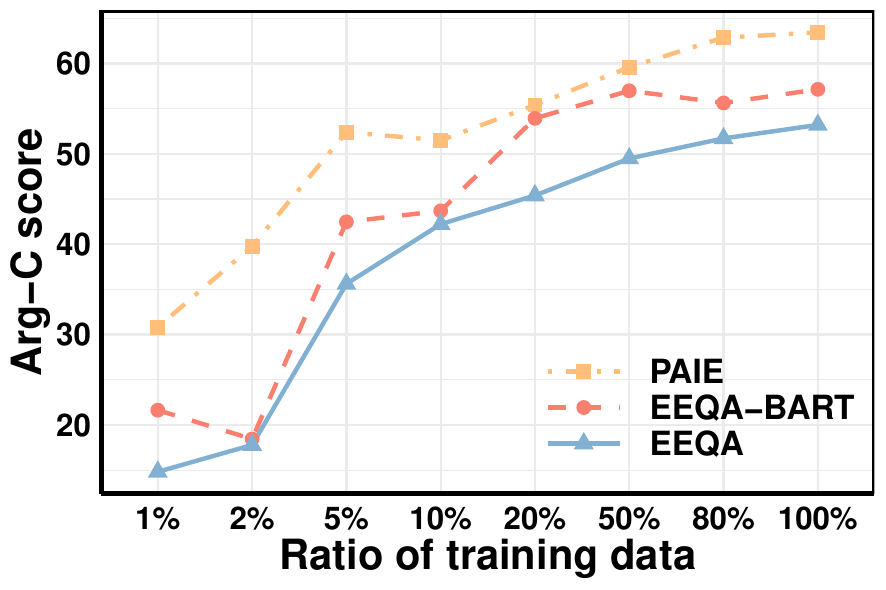}
\end{minipage}
}%
\caption{Arg-C F1 scores w.r.t different training data ratio on three benchmarks.}
\label{fig:fewshot}
\end{figure*}

\noindent{\textbf{Effectiveness}} We also compare the capability of PAIE and EEQA-BART in predicting multiple arguments with the same role on WIKIEVENTS, a dataset containing diverse multi-argument cases. Table~\ref{tab: argnum} shows that PAIE outperforms significantly better than EEQA-BART dealing with such cases. For roles with three and four or more arguments, PAIE gains a definite Arg-C F1 improvement of $9.5\%$ and $26.4\%$, respectively.

\begin{table}
\centering
\small
    \begin{threeparttable}
        \begin{tabular}{B| B B B B}
        \toprule
        \multirow{2}{*}{\textbf{Model}} & \multicolumn{4}{c}{\textbf{WIKIEVENTS Argument Number $n$}} \\
        & $\mathbf{1_{[468]}}$ & $\mathbf{2_{[66]}}$ & $\mathbf{3_{[15]}}$ & $\mathbf{\geq4_{[17]}}$ \\
        \midrule
        \textbf{EEQA-BART} &  $58.0_{(-16)}$& $59.7_{(-3)}$ & $28.6_{(-10)}$ & $10.0_{(-26)}$  \\
        \textbf{PAIE} & \textbf{74.1} & \textbf{62.6} & \textbf{38.1} & \textbf{36.4} \\
        \bottomrule
        \end{tabular}
    \end{threeparttable}
\caption{Arg-C F1 on WIKIEVENTS breakdown by argument number $n$ of one role. The case number is given in the square bracket. }
\label{tab: argnum}
\end{table}

\subsection{Few-shot Setting}
We analyze how PAIE performs under a scenario without sufficient annotations. Fig.~\ref{fig:fewshot} shows the performance of PAIE and two other QA-based baselines with partial training samples on three benchmarks. It demonstrates that \textbf{(1)} PAIE is superior to EEQA-BART and EEQA in almost all settings with different datasets and training data ratios. \textbf{(2)} PAIE especially outperforms QA-based methods in document-level tasks (RAMS and WIKIEVENTS). It achieves comparable F1 scores with EEQA-BART using only about $20\%$ training samples and EEQA using about $10\%$ samples. \textbf{(3)} Along with the decreasing number of training data, the gains become larger than baselines. All observations above indicate that PAIE can better utilize PLMs for few-shot settings.

\subsection{Inference Speed}
\label{sec: inference speed}
Most of the previous sections emphasize the superiority of PAIE from the perspective of accuracy performance. Actually, PAIE also has much better extraction efficiency compared with other approaches.

\begin{table}[t]
    \centering
    \small
    \begin{threeparttable}
    \begin{tabular}{B|BBBBBB}
        \toprule
         \multirow{2}{*}{\textbf{Model}} & \multicolumn{2}{c}{\textbf{ACE05}} & \multicolumn{2}{c}{\textbf{RAMS}} & \multicolumn{2}{c}{\textbf{WIKI}} \\
         & B & L & B & L & B & L \\
         \midrule
         \textbf{BART-Gen} & 5.8 & 12.4 & 33.2 & 54.8 & 19.1 & 29.0 \\
         \textbf{EEQA-BART} & 11.8 & 36.0 & 66.0 & 187.4 & 30.9 & 83.8 \\
         \midrule
         \textbf{PAIE} & \textbf{2.9} & \textbf{8.4} & \textbf{19.0} & \textbf{38.6} & \textbf{8.4} & \textbf{18.3} \\
         \bottomrule
    \end{tabular}
    \end{threeparttable}
    \caption{Inference time (second) for different models on test set of ACE05, RAMS, WIKIEVENTS. Experiments are run on one same NVIDIA-1080Ti GPU.
    }
    \label{tab: speed}
\end{table}

In Table~\ref{tab: speed}, we report the overall inference time for different models. PAIE usually runs 3-4 times faster than EEQA, since it predicts multiple roles simultaneously, while EEQA predicts roles one by one. Other QA-based models are likely to have similar speeds with EEQA due to their sequential prediction structure and training process. Also, as discussed in Section~\ref{subsec:same-role arg}, PAIE is even more advantageous under practical application scenarios since it avoids the heavy threshold tuning.

\section{Conclusion}
We propose a novel model PAIE that effectively and efficiently extracts arguments at both sentence and document levels. We define a new prompt tuning paradigm for extraction tasks, which prompts multiple role knowledge from PLMs via role-specific selectors and joint prompts. Extensive experiments on three standard benchmarks demonstrate our proposed model's effectiveness and the generalization ability in both sentence and document level EAE. We have also conducted ablation studies on the main components, the extractive prompting strategy, and several real scenarios. In the future, we are interested in investigating co-reference as an auxiliary task of EAE and introducing entity information to better determine argument boundaries.

\section*{Acknowledgments}
This study is supported under the RIE2020 Industry Alignment Fund – Industry Collaboration Projects (IAF-ICP) Funding Initiative, as well as cash and in-kind contribution from the industry partner(s). We also thank the KULeuven C1 project Macchina for support.

\bibliography{anthology,custom}
\bibliographystyle{acl_natbib}

\appendix

\section{Dataset and Model}
\label{sec:appendix}

\subsection{Dataset statistics}
\label{ap: dataset}
We evaluate on three common datasets for Event Argument Extraction: ACE05~\cite{doddington-etal-2004-automatic}, RAMS~\cite{ebner-etal-2020-multi} and WIKIEVENTS~\cite{li-etal-2021-document}. 

ACE05 is a joint information extraction dataset providing entity, relation, and event annotation for three languages: English, Chinese, and Arabic. We use its English event annotation for sentence-level EAE tasks. We follow the pre-processing procedure of DyGIE++~\cite{wadden-etal-2019-entity}, which keeps 33 event types and 22 argument roles and collects 4859 arguments in the training set, 605 and 576 in the development and test set respectively.

RAMS is a document-level dataset annotated with 139 event types and 65 semantic roles. Each sample is a 5-sentence document, with trigger word indicating pre-defined event type and its argument scattering among the whole document.

WIKIEVENTS is another document-level dataset providing 246 documents, with 50 event types and 59 argument roles. These documents are collected from English Wikipedia articles that describe real-world events and then follow the reference links to crawl related news articles. They also annotate the coreference links of arguments, while we only use the annotations of their conventional arguments in this task.

Table~\ref{tab: dataset} shows their detailed statistics.

\begin{table}[h]
\small
    \begin{tabular}{ABBB}
    \toprule
     \textbf{Dataset} & \textbf{ACE05} & \textbf{RAMS} & \textbf{WIKIEVENTS}\\
    \midrule
    \textbf{\#Sents} & & & \\
    \textbf{Train} & 17,172 & 7,329 & 5,262 \\
    \textbf{Dev} & 923 & 924 & 378 \\
    \textbf{Test} & 832 & 871 & 492 \\ \hline
    \textbf{\#Args} & & & \\
    \textbf{Train} & 4,859 & 17,026 & 4,552 \\
    \textbf{Dev} & 605 & 2,188 & 428 \\
    \textbf{Test} & 576 & 2,023 & 566 \\ \hline
    \textbf{\#Event} & 33  & 139 & 50 \\
    \textbf{\#Role} & 22 & 65 & 59 \\
    \textbf{\#Arg per Event} & 1.19 & 2.33 & 1.40 \\
    \bottomrule
    \end{tabular}
    \caption{Statistics of datasets.}
    \label{tab: dataset}
\end{table}

\subsection{Details of baseline models}
\label{ap: baselines}
We compare our model with following previous models. (1) \textbf{ONEIE}~\cite{lin-etal-2020-joint}: a joint model extracting entity, relation and event simultaneously. Different from QA-based model, they rely on extracted entities as candidate arguments. (2) \textbf{BART-Gen}~\cite{li-etal-2021-document}: a conditional generation model generating (rather than recognizing the spans) arguments sequentially via a sequence-to-sequence model and prompt. (3) \textbf{EEQA}~\cite{du-cardie-2020-event}: the first Question Answering (QA) based model designed for sentence-level EAE task. (4) \textbf{FEAE}~\cite{wei-etal-2021-trigger}: a QA-based method extended to document-level EAE by considering argument interactions via knowledge distillation. (5) \textbf{DocMRC}~\cite{liu-etal-2021-machine}: another QA-based method with implicit knowledge transfer and explicit data augmentation. 
The implementation details of all baselines are as follow:
\begin{enumerate}
    \item \textbf{FEAE} \cite{wei-etal-2021-trigger}: We report the results from the original paper.
    \item \textbf{DocMRC} \cite{liu-etal-2021-machine}: We report the results from original paper.
    \item \textbf{BART-Gen} \cite{li-etal-2021-document}: For BART-large model, We report the results from origin paper. For BART-base model, we use their code\footnote{ https://github.com/raspberryice/gen-arg} to test its performance on all datasets.
    \item \textbf{EEQA} \cite{du-cardie-2020-event}: We report the results of ACE05 dataset from the origin papers. We use their code\footnote{https://github.com/xinyadu/eeqa} to test its performance on RAMS and WIKIEVENT dataset. In order to generate the question template of these two datasets automatically, we follow the second template setting in \textbf{EEQA}. The question temlpate is \textit{What is the \underline{ROLE} in \underline{TRIGGER WORD}?}.
    \item \textbf{EEQA-BART}: For a fair comparison with our model, we substitute the pre-trained model of EEQA from BERT to BART and call it EEQA-BART. We re-train the model on ACE05, RAMS and WIKIEVENTS datasets. 
    \item \textbf{ONEIE} \cite{lin-etal-2020-joint}: We use their code\footnote{http://blender.cs.illinois.edu/software/oneie/} and re-train the model on ACE05. We don't report its performance on RAMS and WIKIEVENTS because OneIE is a joint model extracting entity, relation and event. However, there is no entity annotation in RAMS and no relation annotation in both RAMS and WIKIEVENTS. Simply dropping the modules related to entity and relation in OneIE achieves abnormally low performance on RAMS and WIKIEVENTS dataset. Therefore it  is somewhat unfair comparing OneIE with our model and other baselines in these two datasets.
    
\end{enumerate}
For the models we re-trained, we keep all other hyper-parameters except learning rates the same with default settings in their original papers. We search the learning rate in [2e-5, 3e-5, 5e-5] and report the test set performance of the model that performs best on the development set.

\subsection{PAIE implementation and training setup}
\label{ap: implementation}
The optimization procedure of PAIE for one sample is shown in the pseudo code~\ref{code: train}. We initialize the weight in encoder-decoder architecture with pre-trained BART models. The contexts in the document-level dataset sometimes exceed the constraint of BART-Encoder and consume prohibitively large memory; thus we add a window centering on the trigger words and only encode the words within the window. We train each large model on single NVIDIA-V100 GPU and each base model on a single NVIDIA-1080Ti GPU. For each setting, we train models with 5 fixed seeds [13, 21, 42, 88, 100] and 3 learning rates [2e-5, 3e-5, 5e-5]. Then we record the test set performance of the model that performs best on the development set for each random seed. The final reported performance is the average value of results w.r.t five different seeds. For model variations mentioned in Section~\ref{sec: model arch}, we only change the input strategy and leave other parts constant. We list other important hyperparameters in Table~\ref{tab:hyperparameters}.
\begin{table}[]
    \centering
    \small
    \begin{threeparttable}
    \begin{tabular}{c|c}
        \toprule
         Hyperparameter & Value \\
         \midrule
         Batch size  & 16 (ACE05) / 4 (Others) \\
         Weight decay  & 0.01 \\
         Training steps & 10000 \\
         Optimizer & AdamW \\
         Adam $\epsilon$ & $1 \times 10^{-8}$ \\
         Adam $\beta_1$/$\beta_2$ & 0.9 / 0.999 \\
         Scheduler & Linear (with 0.1 warmup step) \\
         Max span length & 10 \\
         Max gradient norm & 5.0 \\
         Window size & 250 \\
         Max encoder seq length & 192 (ACE05) / 500 (Others) \\
         Max decoder seq length & 80 \\
         \bottomrule
    \end{tabular}
    \end{threeparttable}
    \caption{Hyperparameters for PAIE}
    \label{tab:hyperparameters}
\end{table}

\DecMargin{10pt}
\begin{algorithm}[h]
\DontPrintSemicolon
\SetAlgoLined
\SetNoFillComment
\LinesNumberedHidden
\SetAlCapHSkip{0pt}
\KwInput{\small X, Pt \tcp*[h]{Context, Prompt tokens}}
\KwData{\small $Y=\{r_{0}:[[s_{0}^{0}, e_{0}^{0}], [s_{0}^{1}, e_{0}^{1}]]\},\{r_{1}:[[s_{1}^{0}, e_{1}^{0}]]\} $} 
$H_{enc},\ H \gets \text{BART}(X)$\;
$\hat P \gets \text{BART-Decoder}(Pt, H_{enc})$\;
$L \gets 0$\tcp*[h]{Initialize loss}\;
\ForEach{\text{role} \textbf{in} $Y$.keys()}{
    \textbf{\upshape Set} $\hat Y_{role}$ \textbf{\upshape to} empty list\;
    \ForEach{$\textup{EMB}_{slot}$ \textbf{\upshape in} $\hat P$.\textup{get\_next}(\text{role})}{
        $\psi \gets \textup{MeanPool}(\text{EMB}_{slot})$\;
        $\psi^{(s)} \gets \psi \circ \boldsymbol{W}^{(s)}$\;
        $\psi^{(e)} \gets \psi \circ \boldsymbol{W}^{(e)}$\;
        $\text{logit}^{(s)} \gets \psi^{(s)} H$\tcp*[h]{cos-sim to H}\;
        $\text{logit}^{(e)} \gets \psi^{(e)} H$\tcp*[h]{cos-sim to H}\;
        \;
        $\hat Y_{role}$.\textup{insert}(\;\ \ \ $\argmaxA\limits_{(i,j) \in L^{2}, i<j}\ \text{logit}^{(s)}(i) + \text{logit}^{(e)}(j)$\;\ )\;
    }
    $Y_{role},\hat Y_{role} \gets \textup{Hungarian}(Y_{role},\hat Y_{role})$ 
    $L \gets L + CrossEntropy(Y_{role},\hat Y_{role})$
}

\caption{Training one sample}\label{code: train}
\end{algorithm}

\subsection{Details of Bipartite Matching loss}
\label{ap: bipartite matching}
We formulate the details of bipartite matching loss in this section. Given $\text{logit}_{k}^{(start)}$ and $\text{logit}_{k}^{(end)}$ from Eq~\ref{eq: logit}, we apply greedy search on predicted start and end position distributions to select the predicted span for each role-specific selector $\theta_k$.
\begin{equation}
 (\hat{s}_k, \hat{e}_k) = \argmaxA_{(i,j) \in L^{2}, i<j}\ \text{logit}_{k}^{(start)}(i) + \text{logit}_{k}^{(end)}(j)
\label{eq: span accurate}
\end{equation}

Denote $y_{r}=[(s_{0},e_{0}), ... , (s_{n},e_{n})]$ as ground truth spans of role $r$ for sample $X$, and $\hat y_{r}=[(\hat s_{0},\hat e_{0}), ... , (\hat s_{m},\hat e_{m})]$ as predicted spans, where $m$ is the number of occurrence of role $r$ in the corresponding prompt. 

\begin{table*}[t]
  \centering
    \begin{tabularx}{\textwidth}{X A A}
    \toprule
     \textbf{\small Example} & \textbf{\small w/ Bipartite} & \textbf{\small w/o Bipartite} \\
    \midrule
    \makecell[Xt]{\small ``We demand that the Security Council ... ," said a spokesman for a \underline{meeting} (\textit{Contact.Meet}) Saturday of \textcolor{blue}{\textbf{Saddam}} and top - level \textcolor{red}{\textbf{officials}} , quoted by media. } & \makecell[lt]{\small \textbf{Entity:} Saddam\\ \small \textbf{Entity:} officials}& \makecell[lt]{\small \textbf{Entity:} Saddam \\ \small \textbf{Entity:} $\emptyset$}\\
    \makecell[Xt]{\small ..., bombing at the world-renowned race, where \textcolor{red}{\textbf{he}} and his brother, \textcolor{blue}{\textbf{Tamerlan}}, 26, \underline{set off} (\textit{Conflict.Attack.DetonateExplode}) two pressure-cooker bombs near... }& \makecell[lt]{ \small \textbf{Attacker:}Tamerlan \\ \small \textbf{Attacker:}he }  &  \makecell[lt]{ \small \textbf{Attacker:}Tamerlan \\ \small \textbf{Attacker:}$\emptyset$ } \\
    
    \bottomrule
    \end{tabularx}
  \caption{Examples from our benchmark datasets. Prediction results for models with/without bipartite matching loss. Argument roles are boldfaced in example sentences, trigger words are underlined, and the event types are in brackets.
  }
  \label{tab: bipartite example}
\end{table*}

With the candidate spans for each role, we define the bipartite matching between the candidates and ground truth annotations as finding the lowest cost of a permutation $\Gamma$ of $N$ elements:
\begin{equation}
 \hat \sigma = \argminA_{\sigma \in \Gamma_{N}} \sum_{k}^{N} L_{1}((s_k,e_k), (\hat s_{\sigma(k)}, \hat e_{\sigma(k)}))
\label{eq: hungarian}
\end{equation}
where $L_{1}((s_k,e_k), (\hat s_{\sigma(k)}, \hat e_{\sigma(k)}))$ represents $L_{1}$-norm between $(s_k,e_k)$ and $(\hat s_{\sigma(k)}, \hat e_{\sigma(k)})$.

We introduce the classical Hungarian algorithm~\cite{kuhn1955hungarian} for efficient optimal assignment. In Eq.\ref{eq: hungarian}, $N$ is chosen to the minimum value between $m$ and $n$. If the number of candidate spans $m$ is larger than the number of ground truth span $n$, we will pad $(0,0)$ representing no arguments to the golden answer set. Otherwise, we only select the optimally matched gold spans for bipartite matching loss calculation.

After finding the optimal assignment $\hat{\sigma}$, we align each ground truth span in $y_r$ and each predicted span in $\hat{y}_r$ according to the matching result and then calculate probabilities where the start/end positions locate about role slot $k$. Note that we use the logit distribution of $\hat{\sigma}(k)$ rather than $k$, which is different from Eq.~\ref{eq: softmax without bipartite} without bipartite matching: 
\begin{align*}
    p_k^{(start)} &= \text{Softmax}(\text{logit}_{\hat{\sigma}(k)}^{(start)}) \\
    p_k^{(end)} &= \text{Softmax}(\text{logit}_{\hat{\sigma}(k)}^{(end)})
\label{eq: softmax with bipa}
\numberthis
\end{align*}

Given $p_k^{(start)}$ and $p_k^{(end)}$ obtained by Eq.~\ref{eq: softmax with bipa}, we follow the same loss function in Eq.~\ref{eq: loss} during training process. The bipartite matching is only applied in training. For inference, the model will output all non-zero spans with corresponding argument roles as predictions.

\begin{table*}[t]
\small
\centering
\begin{threeparttable}
  \begin{tabular}{c|c}
    \toprule
    \textbf{Prompt Type} & \textbf{Prompt Example} \\
    \midrule
    \makecell{Question Answering Prompt \\ \cite{du-cardie-2020-event}} & \makecell{Who is the Victor in the Conflict.defeat event? \\ What is the ConflictOrElection in the Conflict.defeat event? \\ Where is the Place in the Conflict.defeat event?} \\[+0.5ex]
    \midrule
    \makecell{Conditional Generation Prompt \\ \cite{li-etal-2021-document}}
    & <arg1> defeated <arg2> conflict at <arg3> place \\[+0.5ex]
    \midrule
    Manual Template (Ours) & \underline{Victor} ( and \underline{Victor} ) defeated in \underline{ConflictOrElection} at \underline{Place} ( and \underline{Place} ) \\[+0.5ex]
    \hline & \\[-2ex]
    Concatenation Template (Ours) & \underline{Victor} ( \underline{Victor} ) \underline{ConflictOrElection} \underline{Place} ( \underline{Place} )\\[+0.5ex]
    \midrule
    Soft Prompt (Ours) & \makecell{<Vic\_left0> \underline{Victor} <Vic\_right0> ( <Vic\_left0> \underline{Victor} <Vic\_right0> ) \\ Defeated <Conf\_left0> \underline{ConflictOrElection} <Conf\_right0> \\ <Place\_left0> \underline{Place} <Place\_right0> ( <Place\_left0> \underline{Place} <Place\_right0> )} \\
    \bottomrule
  \end{tabular}
    \caption{Example prompts about Event type \textit{Conflict.Defeat.Unspecified} in WIKIEVENTS dataset. Angle brackets in conditional generation prompt denote the content to be filled during the decoding stage. Angle brackets in soft prompt represents pseudo tokens connecting different slots. Underlined words in the last three rows denote role slots, and brackets include roles with multiple arguments.}
  \label{tab: prompt template1}
  \end{threeparttable}
\end{table*}

\subsection{Further analysis of Bipartite Matching}
\label{sec: bipartite discuss}
Ablation studies have validated the effectiveness of bipartite matching loss. In our settings, bipartite matching loss focuses on multiple arguments of the same role and reassigns the predicted arguments in each prompt slot. Since slots in our joint prompts usually entail different semantic meanings and matching preferences, even they are about the same roles, the permutation-invariance property of bipartite matching assures a global optimization of these arguments.

Such optimization especially makes sense when the arguments in context have subtle semantic distinction, and such distinction can not merely be captured by sequential order. Simple argument enumerations, for example, do not satisfy the condition mentioned above, while contexts with different syntactic structures are more likely to satisfy it. We consider such an instance when the prompt is in active voice \textit{\underline{Person} teaches \underline{Person}}, but the context sentence is in passive voice \textit{Peter is taught by John}.\footnote{This prompt and context sentence are imaginary and do not relate to any events/samples in three benchmarks. We use it just for the convenience of discussion.} The position-wise assignment will be likely to assign \textit{Peter} to the first \textit{\underline{Person}} slot and \textit{John} for the second. It is not semantically correct, although they are treated as correct in evaluation. This supervision signal will force the model downgrading to extract arguments by position order during the training process. Such a model is hard to be voice-aware (and also insensitive to capture other syntactic structures) and tends to misidentify multi-argument data during inference, as shown in Table~\ref{tab: bipartite example}.

However, we also acknowledge the improvement from bipartite matching loss is somewhat not significant and robust when compared with other contributions in our paper. We attribute it to the following points: \textbf{(1)} existing datasets are not designed especially for evaluating the multi-arguments problem, only $8.9\%$ samples in ACE05, $6.1\%$ in RAMS and $10.9\%$ in WIKIEVENTS facing it. \textbf{(2)} Even in limited cases about multi-arguments, related arguments are usually simply enumerated and do not require complex analysis and matching about implicit structure. Thus we expect a large-scale dataset with more multi-arguments and diverse narrative styles in the future, and we believe the bipartite matching loss will bring more significant improvement in it.
\section{Prompt Examples}
\label{subsec: prompt example}
We compare our prompt with others used in EAE task in Table~\ref{tab: prompt template1}. The first row gives a standard QA-based template, and the second row shows a standard prompt in the generation paradigm. Row 3-5 show our three types of joint prompts respectively. 

We further show ten manual template examples of each dataset at Table \ref{tab: more prompt template}. The complete version of different types of prompts is available in our codebase.

\begin{table*}[t]
\small
\centering
\begin{threeparttable}
  \begin{tabular}{D|c|D}
    \toprule
    \textbf{Dataset} & \textbf{Event Type} & \textbf{Natural Lanugage Prompt} \\
    \midrule
     \multirow{15}{*}{ACE05}{\centering} & Movement.Transport & \makecell[c]{\underline{Agent} (and \underline{Agent}) transported \underline{Artifact} (and \underline{Artifact})  in \underline{Vehicle} \\ (and \underline{Vehicle})  cost  \underline{Price} from \underline{Origin} place (and \underline{Origin} place)  \\ to \underline{Destination} place (and \underline{Destination} place)}   \\
    \cline{2-3}
    & Justice.Arrest-Jail & \makecell[c]{ \underline{Agent} (and \underline{Agent}) arrested \underline{Person} (and \underline{Person}) \\ at \underline{Place} (and \underline{Place}) for \underline{Crime}} \\ 
    \cline{2-3}
    & Justice.Execute & \makecell[c]{\underline{Agent} (and \underline{Agent}) executed \underline{Person} at \underline{Place} (and \underline{Place}) for \underline{Crime} } \\
    \cline{2-3}
    & Conflict.Attack & \makecell[c]{\underline{Attacker} (and \underline{Attacker})  attacked \underline{Target} (and \underline{Target}) \\ hurting \underline{Victims} using \underline{Instrument} (and \underline{Instrument}) at \underline{Place} (and \underline{Place})} \\
    \cline{2-3}
    & Contact.Meet& \underline{Entity} (and \underline{Entity}) met with \underline{Entity} (and \underline{Entity}) at \underline{Place} (and \underline{Place}) \\
    \cline{2-3}
    & Conflict.Demonstrate & \makecell[c]{\underline{Entity} (and \underline{Entity}) demonstrated at \underline{Place} (and \underline{Place})} \\ 
    \cline{2-3}
    & Transaction.Transfer-Ownership & \makecell[c]{\underline{Seller} gave \underline{Buyer} ( and \underline{Buyer}, \underline{Buyer}, \underline{Buyer}, \underline{Buyer}, \underline{Buyer}, \underline{Buyer} ) to \\ \underline{Beneficiary} ( and \underline{Beneficiary}, \underline{Beneficiary} ) for the benefit of \\ \underline{Artifact} ( and \underline{Artifact}, \underline{Artifact} ) cost \underline{Price} at \underline{Place} ( and \underline{Place}, \underline{Place)} } \\ 
    \cline{2-3}
    & Transaction.Transfer-Money & \makecell[c]{\underline{Giver} (and \underline{Giver}) gave \underline{Money} to \underline{Recipient} (and \underline{Recipient}) \\for the benefit of \underline{Beneficiary} (and \underline{Beneficiary}) at \underline{Place} (and \underline{Place}) } \\ 
    \cline{2-3}
    & Life.Be-Born & \underline{Person} (and \underline{Person}) was born at \underline{Place} (and \underline{Place}) \\
    \cline{2-3}
    & Life.Marry & \underline{Person} married \underline{Person} at \underline{Place} (and \underline{Place}) \\
    \midrule
    \multirow{18}{*}{RAMS} & \makecell[c]{life.injure.\\illnessdegradationphysical} & \makecell[c]{\underline{Victim} person has some physical degradation \\ from \underline{Medicalissue} imposed by \underline{Injurer} at \underline{Place}} \\
    \cline{2-3}
     & \makecell[c]{artifactexistence.\\damagedestroy.destroy} & \underline{Destroyer} destroyed \underline{Artifact} using \underline{Instrument} in \underline{Place} \\
    \cline{2-3}
     & conflict.yield.surrender & \underline{Surrenderer} surrendered to \underline{Recipient} at \underline{Place} \\
    \cline{2-3}
     & conflict.yield.retreat & \underline{Retreater} retreated from \underline{Origin} place to \underline{Destination} place \\
    \cline{2-3}
     & \makecell[c]{contact.commandorder.\\correspondence} & \makecell{\underline{Communicator} communicated remotely \\ with \underline{Recipient} about \underline{Topic} at \underline{Place}} \\
    \cline{2-3}
     & \makecell[c]{government.agreements.\\rejectagreementcontractceasefire} & \makecell{\underline{Rejecternullifier} rejected or nullified an agreement \\ with \underline{Otherparticipant} in \underline{Place}} \\
    \cline{2-3}
     & \makecell[c]{government.vote.\\violationspreventvote} & \makecell{\underline{Preventer} prevented \underline{Voter} from voting \\ for \underline{Candidate} on ballot in \underline{Place}} \\
    \cline{2-3}
     & \makecell[c]{inspection.sensoryobserve.\\physicalinvestigateinspect} & \underline{Inspector} inspected \underline{Inspectedentity} in \underline{Place} \\
    \cline{2-3}
     & \makecell[c]{manufacture.artifact.\\createintellectualproperty} & \makecell{\underline{Manufacturer} manufactured or created or produced \\ \underline{Artifact} using \underline{Instrument} at \underline{Place}} \\
    \cline{2-3}
     & \makecell[c]{life.injure.\\illnessdegredationsickness} & \makecell{\underline{Victim} has disease sickness or illness at \underline{Place}, \\ deliberately infected by \underline{Injurer}} \\
    \midrule
    \multirow{20}{*}{\shortstack{WIKI-\\EVENTS}} & \makecell[c]{ArtifactExistence.\\ManufactureAssemble} & \makecell[c]{\underline{ManufacturerAssembler} (and \underline{ManufacturerAssembler}) \\ manufactured or assembled or produced \underline{Artifact} (and \underline{Artifact}) \\ from  \underline{Components} (and \underline{Components}) using \\ \underline{Instrument} (and \underline{Instrument}) at \underline{Place} (and \underline{Place})} \\
    \cline{2-3}
     & Conflict.Demonstrate & \makecell[c]{\underline{Demonstrator} was in a demonstration for \underline{Topic} \\ with \underline{VisualDisplay} against \underline{Target} at \underline{Place}, \\ with potential involvement of \underline{Regulator} police or military} \\
    \cline{2-3}
     & \makecell[c]{Cognitive.Inspection.\\SensoryObserve} & \makecell[c]{ \underline{Observer} (and \underline{Observer}) observed  \underline{ObservedEntity} \\ (and \underline{ObservedEntity}) using \underline{Instrument} \\ (and \underline{Instrument}) in \underline{Place} (and \underline{Place})} \\
    \cline{2-3}
     & \makecell[c]{Cognitive.\\TeachingTrainingLearning} & \makecell[c]{\underline{TeacherTrainer} (and \underline{TeacherTrainer}) taught \\ \underline{FieldOfKnowledge} (and \underline{FieldOfKnowledge}) \\to \underline{Learner} (and \underline{Learner}) using \underline{Means} (and \underline{Means}) \\at \underline{Institution} (and \underline{Institution}) in \underline{Place} (and \underline{Place})}\\
    \cline{2-3}
     & \makecell[c]{Control.ImpedeInterfereWith} & \makecell[c]{ \underline{Impeder} (and \underline{Impeder}) impeded or interfered \\ with \underline{ImpededEvent} at \underline{Place} (and \underline{Place})}\\
    \cline{2-3}
     & \makecell[c]{Transaction.Donation} & \makecell[c]{\underline{Giver} gave \underline{ArtifactMoney} to \underline{Recipient} (and \underline{Recipient}) for \\ the benefit of \underline{Beneficiary} (and \underline{Beneficiary}) at \underline{Place} (and \underline{Place})}\\
    \cline{2-3}
     & Disaster.DiseaseOutbreak & \makecell[c]{\underline{Disease} (and \underline{Disease}) broke out among \underline{Victim} (and \underline{Victim}) \\ or population at \underline{Place} (and \underline{Place})} \\
    \cline{2-3}
     & Justice.TrialHearing & \makecell[c]{\underline{Prosecutor} tried \underline{Defendant} (and \underline{Defendant}) before \underline{JudgeCourt} \\ for \underline{Crime} (and \underline{Crime}) in \underline{Place} (and \underline{Place})} \\
    \cline{2-3}
     & Medical.Vaccinate & \makecell{\underline{Treater} vaccinated \underline{Patient} via \underline{VaccineMethod} \\ for \underline{VaccineTarget} at \underline{Place} (and \underline{Place})} \\
    \cline{2-3}
     & Personnel.StartPosition & \makecell[c]{\underline{Employee} started working in \underline{Position} at \underline{PlaceOfEmployment} \\ organization in \underline{Place} (and \underline{Place})} \\
    \bottomrule
  \end{tabular}
    \caption{Example manual templates used in our work. Underlined words denote role slots, and slots in brackets denote repetitive ones designed for multi-arguments of the same roles.}
  \label{tab: more prompt template}
  \end{threeparttable}
\end{table*}

\end{document}